\documentclass[sn-mathphys]{sn-jnl}

\jyear{2021}%
\usepackage{bm}
\theoremstyle{thmstyleone}%
\newtheorem{theorem}{Theorem}
\newtheorem{proposition}[theorem]{Proposition}%

\theoremstyle{thmstyletwo}%

\theoremstyle{thmstylethree}%

\begin{document}

\title[Robust Newsvendor Problem in Global Market]{Robust Newsvendor Problem in Global Market: Stable Operation Strategy for a Two-Market Stochastic System}


\author[1]{\fnm{Xiaoli} \sur{Yan}}\email{20170602013t@cqu.edu.cn}


\affil[1]{\orgdiv{School of Economics and Business Administration}, \orgname{Chongqing University}, \city{Chongqing}, \postcode{400044}, \country{China}}


\abstract{The global markets provide enterprises with selling opportunities and challenges in stabilizing operational strategies. From the perspective of production management, it is important to improve the profitability of an enterprise by exploiting the different timing of the selling season in different markets to develop an operational strategy that is optimized and configured on a global scale. This paper examines the above issue with an insightful model of selling the product to two markets (a primary and a secondary market) with multiple risks of changes in market environment and nonoverlapping selling seasons. We refer to this problem as the ``global robust newsvendor" problem. For the above two market stochastic systems, we provide closed-form solutions of the optimal operation strategy for demand-independent and demand-related scenarios, respectively. The closed-form solutions fully reflect the influence of the relationship between supply and demand on strategy selection. We find that the demand correlation and the lack of demand information will not substantially affect the operation strategy, and the industrial chain and supply chain of the enterprise remain stable. However, the reduction of inter-market tariffs or logistics costs will cause changes, and the existence of the secondary market will lead to more capacity planning in the primary market. In addition, our model explicitly considers the impact of exchange rate uncertainty on operating strategies. }




\maketitle

\section{Introduction}\label{sec1}

The global markets offer to the enterprise more selling opportunities and choice of production locations (Kogut and Kulatilaka 1994; Kouvelis and Gutierrez 1997). For example, Elon Musk once stated that Tesla plans to build factories on every continent and currently operates seven. Among them, Tesla's German factory began construction in the first quarter of 2020, and plans to start production in July 2021 (the initial production of Model Y and later production of Model 3). During this period, due to the surge in demand for Model Y in California, the production focus of the California plant was shifted to Model Y, which caused Model 3 to still need to be imported before localized production in Europe. On November 26, 2020, the Shanghai plant exported more than 3,000 Model 3s to Europe for sale. It can be seen that Tesla has the ability to deploy production and export transshipment on a global scale, and has been actively seeking a global strategy to optimize its operations. Another important advantage is the ability to exploit different selling seasons at various markets. For example, there is a six-month time lag between the product life cycles of various models in the North American and European markets, and North American manufacturers can dispose the remaining quantities of outdated sports shoes to the European market; American clothing manufacturers can also sell the remaining summer fashion to the Australian market, which is about to usher in summer. This paper assesses the enterprise's transnational operation strategy for optimization and configuration in global markets with non-overlapping geographic location and selling seasons.
    
For global enterprises facing overseas markets, their products flow across national boundaries, which also leads to the uncertainty and complexity of the risks of global environmental changes. The currency exchange rate in production input and the uncertainty of market demand are two complex factors in the global supply chain environment. Exchange rate risk especially affects the basic economic conditions of enterprises through input costs, sales prices, and sales quantity (Ding et al. 2017), and fluctuations of 1\% per day or 20\% per year are not surprising (Dornier et al. 1998). In addition, in today's deeply integrated development of economic globalization, countries are becoming increasingly interconnected and interdependent. For example, the ``Regional Comprehensive Economic Partnership Agreement” (RCEP) signed on November 15, 2020, the trade in goods and rules of origin in the agreement will further promote direct foreign investment among member states and provide great convenience for the establishment of a transnational trading system for enterprises. Thus, the existence of multiple risks (exchange rate, demand, market linkages, etc.) raises a critical research question: what is the optimal mechanism for enterprises to maintain a stable joint operation strategy (production and transshipment strategy) under the multiple risks of changes in the external environment of the global market?
      
Scarf (1958) robust newsvendor model and its variations are the basis of most existing operating strategy literature, which is an unconventional method to solve operational management models with model uncertainty. Traditional operation management models usually require complete model information, but in practice, model ambiguity or uncertainty is ubiquitous (Lu and Shen 2020). In the past few decades, fueled by both the volatile business environment and significant advances in optimization theory, robust methods achieve explosive growth in operations management (see, e.g., Bertsimas et al. 2018; Chen et al. 2019; Delage and Ye 2010; He et al. 2020; Lim and Wang 2017; Mamani et al. 2017; Zhang et al. 2016). It is a promising research direction to develop a robust method for operational decision-making problems in a scarce data environment (Govindan and Cheng 2018;  Govindan et al. 2017).
  
Designing a stable and feasible operation strategy for the above problems in the face of complex and uncertain environments is of great significance from both practical and theoretical perspectives, and provides motivation for our research. However, although the presence of geographically dispersed product markets creates profitable opportunities for enterprises in multiple secondary markets, the problem of enterprises managers' operational strategy becomes more complicated, and beyond the scope of robust newsvendor models that have so far appeared in the literature. Most of the existing robust newsvendor models (including the classic newsvendor model) only focus on the uncertainty of a single factor, and there are few literatures on transnational operation strategies under multiple risks (see, e.g., Fu et al. 2018; Sodhi 2005; Tibben-Lembke 2004). To the best of our knowledge, the model discussed in this study is the first attempt to introduce robust newsvendor ideas into the transnational operation problem of a two-market stochastic system. 
 
Our research has two key features that can be used to distinguish previous studies. On the one hand, we only know the first and second moments of random demand, and use cross moments to characterize the correlation between markets. However, previous studies usually assume that the distribution of demand information is known, and the impact of market relevance on the operating strategy is not given (see, e.g., Kouvelis and Gutierrez 1997; Petruzzi and Dada 2001; Hu et al. 2007; Feng et al. 2019; Rudi et al. 2001; Shao et al. 2011). On the other hand, the global market we studied is a two-market stochastic system with two production bases. The focus of our research is on how enterprises deploy production capacity and allocate resources between these two production bases. It is different from the previous discussion of global operation issues that are almost algorithm-based (see, e.g., Van Landeghem and Vanmaele 2002; Peidro et al. 2010). Specifically, this paper provides a more interpretable and stable operation strategy for enterprises in the uncertain environment of dual markets with scarce data. That is, when to produce in two markets, how many are produced, when to transfer to the secondary market, and what is the quantity of transhipment.
   
Our analysis provides several important managerial implications:
\begin{itemize}  
\item Through the global robust newsvendor method, we provide the basic analysis framework of the enterprise operating in two random markets with two production bases. 
\item Demand correlation and lack of demand information will maintain the stability of the original operating strategy, while factors such as inter-market friction costs (tariffs, logistics costs, etc.), exchange rates, and selling prices will affect the enterprise's operational strategy. 
\item The production strategy without transshipment is subject only to demand risk and remains independent and symmetric. The production strategy at the time of transshipment is also associated with exchange rate risk, but will lose its independence.  
\item The existence of foreign markets does provide enterprises with larger product markets and better arbitrage opportunities. That is, whether an enterprise invests in production in a foreign production center or not, it will be more motivated to plan more production capacity in the domestic market.
\end{itemize}
         
The remainder of this paper is organized as follows. Sect.~\ref{sec2} reviews the relevant literature. Sect.~\ref{sec3} introduces the research questions. Sect.~\ref{sec4} presents the closed-form multinational operation strategy of the enterprise for demand-independent, and examines its robustness. Sect.~\ref{sec5} reproduces the corresponding management insights from the perspective of enhanced market linkages (demand-related). Sect.~\ref{sec6} concludes the paper. All proofs are relegated to the Appendix.

\section{Literature review}\label{sec2}
\citet{Andersson2013}

The current work is closely related to three streams of literature: transnational supply chain, inventory transshipment, and robust newsvendor problem. To highlight our contributions, we review the existing studies on these three streams and point out the differences between the current work and previous works (shown in Table~\ref{tab1}).

\begin{table}[h]
\begin{center}
\begin{minipage}{400pt}
\caption{Brief summary of issues and settings considered in the literature}\label{tab1}%
\begin{tabular}{@{}lclcccc @{}}
\toprule
Author(s) & Model & Random variable & Transfer  &  Interna & Correla  & Market  \\ 
  & setting &  & decision & -tional & -tion  &  number  \\
\toprule 
Peidro et al. (2010)  & FLP  & demand \& process &  $\checkmark$  & $-$ & $-$ &1\\
 &   &\& supply &  &  &  &  \\
 Mamani et al. (2017)  & RO  &  N-dimensional  &  $-$  &  $-$  & $\checkmark$  &  1  \\
 &  & demand  &  &  & &  \\
  Arcelus et al. (2013) &  CNV & demand  & $-$ &  $\checkmark$   &  $-$  & 1  \\
 Roghanian & BLP & demand \& capacity  & $-$ & $-$ & $-$ & 1 \\
 et al. (2007) &  & \& available  &  &  &  &   \\
 Kouvelis and  & CNV & demand \& demand & $\checkmark$ &  $\checkmark$ & $-$ &  2\\
 Gutierrez (1997) &  & \& exchange rate &  &  &  &  \\
 Zhang et al. (2016)  & RO & demand  & $-$  &  $-$  & $\checkmark$   &  1  \\
 Petruzzi and  & CNV & demand \& demand  &  $\checkmark$  & $\checkmark$  & $\checkmark$  & 2 \\
 Dada (2001)  &   &   &   &   &   &    \\
 Chen et al. (2019) & RO  & N-dimensional  & $-$  &  $-$  & $-$  & 1   \\
&  & demand  &  &  & &  \\
 Shao et al. (2011) & RO  &demand \& demand  & $\checkmark$   &  $-$  &  $-$  & 2   \\
 Huang et al. (2014) & CNV & demand  & $-$ & $-$ &  $\checkmark$ &  2\\
 Zimmer et al. (2017) & FAHP & $-$ & $-$ & $\checkmark$ & $-$  &  $>2$\\
 Naderi et al. (2020) &MILG  &$-$   & $\checkmark$  & $-$  & $-$ & $>2$ \\
Feng et al. (2019) & CNV & demand \& demand & $\checkmark$ & $-$ &$-$  &  2\\
 Fu et al. (2018) & RNV & demand \& price  & $-$ & $-$ & $\checkmark$  &1  \\
 This study & RNV & demand \& demand & $\checkmark$ & $\checkmark$  & $\checkmark$  & 2 \\
 &  & \& exchange rate&  &  & &  \\
\botrule
\end{tabular}
\footnotetext{FLP: Fuzzy linear programming; BLP: bi-level linear programming; FAHP: Fuzzy analytical hierarchy process; MILG: Mixed integer linear programming; RO: Robust optimization(design algorithm); CNV: Classic newsvendor model; RNV: Robust newsvendor model.}
\end{minipage}
\end{center}
\end{table}

\subsection{Transnational Supply Chain}\label{subsec2}

Due to the intensification of globalization and international competition, transnational supply chain management becomes crucial. In academia, scholars study various aspects of transnational supply chains, such as supplier selection and logistics network optimization (Vidal and Goetschalckx 2001; Zimmer et al. 2017), environmental protection and human rights issues (Ras and Vermeulen 2009; Martin-Ortega 2017), and factors affecting the performance of transnational supply chains (Arcelus et al. 2013). Bok et al. (2000) provide a multi-period supply chain optimization model. Wilkinson et al. (1996) propose a method to integrate production and distribution in a multi-site facility. Jackson and Grossmann (2003) study a time decomposition scheme based on Lagrangian decomposition. None of the above models consider the uncertainty or risk in the supply chain planning process. However, transnational supply chains face more uncertainties and uncontrollable factors than domestic supply chains, making transnational supply chains more susceptible to multiple risks (Meixell and Gargeya 2005). Therefore, the issue discussed in this paper explicitly considers the risk of demand and exchange rate fluctuations.
        
In addition, many methods are proposed in the existing literature to deal quantitatively with uncertainties in design, strategy, and scheduling problems. Subrahmanyam et al. (1994) use a scenario-based approach. However, although the scenario-based approach provides a direct way to implicitly consider uncertainty, the scale of the problem will increase exponentially as the number of solutions increases. Other methods to solve uncertainty, such as Monte Carlo simulation (Van Landeghem and Vanmaele 2002) and fuzzy linear programming (Peidro et al. 2010; Roghanian et al. 2007) in the framework of project planning. When considering the uncertainty risk of the supply chain, these studies are almost all based on algorithms. The difference in this paper is that it shows some conditions after weighing the main factors of the market, and these conditions are important. It is more interpretable than pure algorithmic design, such as enterprises can know specifically when (not) to produce and when (not) to transfer after effective supply and demand matching. In addition, the closed-form solution of the optimal strategy that we provide for enterprises, combined with the approximate optimality of the solution, is basically more guiding significance to practice and management than the research perspective of the classic newsvendor. For other important work on the strategic planning and management of global production and distribution networks can refer to Govindan et al. (2017) and Meixell and Gargeya (2005).
 
\subsection{Inventory Transshipment}\label{subsec3}

The research of Tibben-Lembken (2004) shows that retail enterprises usually fail to sell about 5\% to 20\% of their products in the primary market. In fact, in real business operation, inventory transfer strategy is widely used in such retail industries as automobiles, computers, clothing, etc., and has become a routine process of operation and management for some enterprises.
  
The research on inventory transfer originated from Krishnan and Rao (1965), who established a product transfer model among multiple retailers with independent demand and centralized control. After that, Tagaras (1989), Herer and Rashit (1999) carry out further research based on their model framework. In addition, the emergency transshipment strategy is first proposed by Lee (1987). Subsequently, Axsäte (1990) assumes that the replenishment time obeys an exponential distribution, and improves Lee (1987)'s model by using the single commodity queuing theory.
    
However, the markets in existing studies are almost overlapped (e.g., Arcelus et al. 2013), because some consumers are co-consumers of the two markets, and demand occurs simultaneously during the selling season. Although our research question also takes inventory transfer into account, the two markets we consider do nonoverlap. In addition, common transshipment only occurs when there is surplus inventory in one market and a shortage in another (Dong and Rudi 2007). However, the transshipment involved in this paper can directly transfer surplus inventory from one market to another market, and then the another market can make production decisions based on demand and transshipment volume. As far as we know, there are very few such studies (demand in each market occurs in nonoverlapping time periods) at present. Available existing studies such as Kouvelis and Gutierrez (1997), Huang et al. (2014), and Petruzzi and Dada (2001).

\subsection{Robust Newsvendor Problem}\label{subsec4}

In the design method of the operation strategy, Govindan et al. (2017) and Govindan and Cheng (2018) point out that we should pay more attention to the robust method compared with fuzzy programming and stochastic programming. Scarf (1958) first studied the robust newsvendor problem that only mean and variance of the demand distribution known. The idea is to find the worst distribution form among all possible distributions of demand, and transform the objective function to maximize the expected profit under the worst distribution, thereby obtaining a closed-form robust ordering rule. Later, Gallego and Moon (1993) re-solve the robust newsvendor model proposed by Scarf in a concise proof, and apply it to several variants.
  
Kouvelis and Gutierrez (1997) is one of the literatures closest to the research in this paper. They study the “global newsvendor" problem of selling “style goods” to two markets without overlapping time and location.  Subsequently, Petruzzi and Dada (2001) incorporate updated information on retail pricing and demand distribution between the two markets into the problem of Kouvelis and Gutierrez (1997). However, the premise of these studies is based on the specific distribution of demand, but the actual decision-making environment is often uncertain, and it is difficult to obtain an accurate distribution. At the same time, the distribution itself is to portray uncertain information, it actually has a certain degree of assumptions, and changes in the business environment will change the demand pattern of the market, so the distribution of demand will also change over time (Levi et al. 2015). Instead, this paper focuses on global operational strategy issues with only partial information about the demands of the two markets. Our model allows us to obtain a closed-form operational strategy that quantifies the specific operating conditions and size of an enterprise in an uncertain environment. This paper can be a supplement to the study of Kouvelis and Gutierrez (1997).
          
The literature on the distributionally robust method that only knows the first and second moments of two random variables has only recently appeared. From the perspective of modeling development, the work of Fu et al. (2018) is the most influential. They develop a distributionally robust Stackelberg game model that solves the problem of how the decentralized supply chain should use minimum demand and price information to set up a profit sharing contract. The proposal of the method based on two random variables breaks through and extends the traditional robust newsvendor model of single random variables. The research method in this paper is inspired by Fu et al. (2018), and discusses the design issues of operation strategy under the uncertain demand of the two markets. Interestingly, we find that the correlation between demands does not affect the enterprise's operational strategy, which saves enterprise planners the trouble of considering many existing problems in operational management.

\section{Problem description}\label{sec3}
\begin{figure}[h]%
\centering
\includegraphics[width=0.9\textwidth]{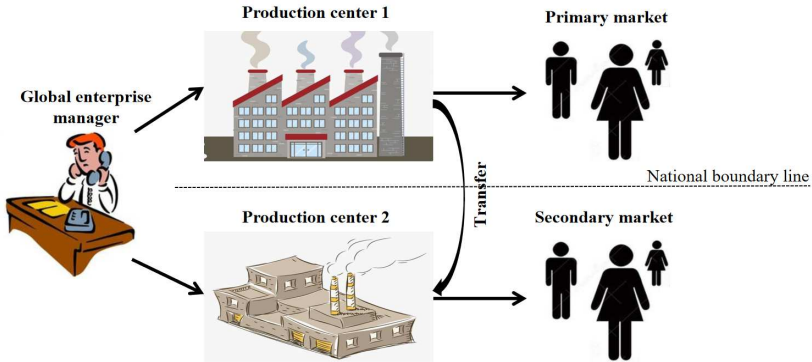}
\caption{The two-market stochastic system for the enterprise}\label{fig1}
\end{figure}

Consider an enterprise that sells products in two random markets with non-overlapping selling seasons. The enterprise has two production centers for the item, one for each market. In addition to the production strategy of the two production centers, the enterprise also needs to consider whether to transfer part or all of the surplus inventory of one market to another market (transshipment strategy), so as to better balance production and demand (see Figure~\ref{fig1}).  

For the convenience of presentation, the market of export products is referred to as the ``primary market", and  the other market is called the ``secondary market". Let $D_{i},i=1,2,$ be the random variable describing the demand for products in market $i$, and their mean, variance, and covariance information can be estimated based on expert judgment or survey opinions. The index 1 is related to the primary market, and 2 to the secondary one. To explore the enterprise's stable operation strategy, two scenarios of demand independence and demand correlation are discussed. The goal of the enterprise is to maximize the expected total profit generated by the two markets in these two situations.

\begin{figure}[h]%
\centering
\includegraphics[width=0.99\textwidth]{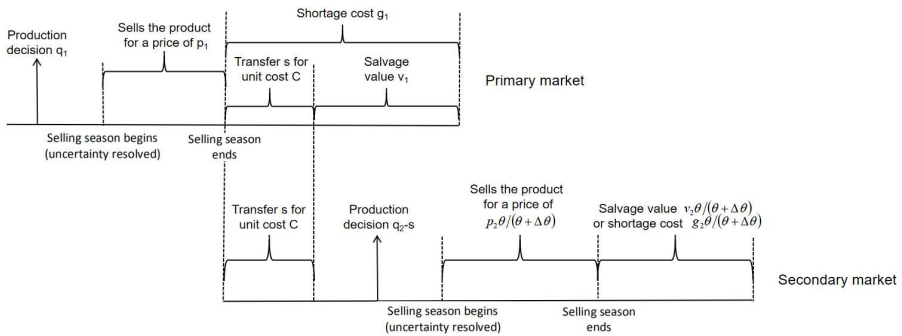}
\caption{The operation process of the enterprise in the multinational market}\label{fig2}
\end{figure}

Figure~\ref{fig2} shows the operation process of enterprises deploying production and transshipment in multinational markets. First, make a decision on the production strategy $q_{1}$ in the primary market, and then make a decision on the transfer strategy $s$ and the production strategy $q_{2}-s$ in the secondary market. Let $c_{i}$ be the production cost at center $i$ in the currency of market $i$. The selling price of the product at the end of the selling season (salvage value) in market $i$ is $v_{i}$ (in the currency of market $i$), and the unit shortage cost in market $i$ is $g_{i}$ (in the currency of market $i$) and it accounts for the opportunity cost on lost sales. The transportation cost of products from market 1 to market 2 is $C$ (in the currency of market 1) and it explains the cost of entering the secondary market, including tariffs, logistics costs, etc. 

To describe the decision-making situation in reality, we assume $p_{i}\geq g_{i}\geq c_{i} \geq v_{i} $ $(i=1,2)$ in the following analysis. Among them, $c_{i} \geq v_{i} $ $(i=1,2)$ is to ensure that the enterprise has the motivation to produce and sell the product, and $g_{i}\geq c_{i}$ $(i=1,2)$ is made to avoid the optimal situation not to sell any unit. In addition, we also assume $g_{1}\geq c_{2}$ to eliminate the trivial situation where the primary market manager suppresses sales in the primary market in order to sell in the secondary market.
    
In this paper, we assume that all price and cost parameters are known and can be reasonably estimated by the enterprise before making corresponding decisions (Lin and Ng 2011). However, it is difficult to understand the demands of the market, which may be highly uncertain. In fact, the random factor that affects enterprise operation is not only the market demand, but also at least the risk of exchange rate fluctuations. Exchange rate fluctuations affect some relevant price factors of operational decisions, such as selling price $p_{2}$, production cost $c_{2}$, shortage cost $g_{2}$ and excess cost $v_{2}$ in the secondary market. Our model takes into account the risk of exchange rate uncertainty. The real exchange rate between the primary market and the secondary market is represented by $\theta$ (i.e., $\theta$ units of the primary market currency correspond to one currency unit of the secondary market). From the setting of the production strategy in the secondary market to the actual production period, the exchange rate of the two countries may change, which may cause exchange rate fluctuation risks $\Delta \theta$. In addition, we also assume that the enterprise alone bears the risks related to exchange rate, and that the tariffs and logistics costs involved in the export of products are included in the transshipment cost $C$. 

\section{Two-market single product global robust newsvendor problem under independent demand}\label{sec4}

In the rest of the paper, we formally derive the main results obtained in this paper. By establishing a global robust newsvendor model in a centralized decision-making environment, section~\ref{subsec2} presents the optimal transshipment and production strategy for the secondary market production center; section~\ref{subsec3} provides the optimal production strategy for the primary market production center; then section~\ref{subsec4} conducts a robustness test on the proposed optimal production strategy.
  
\subsection{Optimal transshipment and production strategies on the secondary market}\label{subsec2}

This section analyzes the decision-making problems faced by enterprises in the secondary market, that is, how much to transfer from the primary market, how much to produce in the secondary market production center, and what are the respective conditions.
  
Let $x$ denote the remaining inventory at the end of the selling season at the primary market ($s\leq x$), then the optimal profit function $\Pi_{2 }$ of the enterprise in the secondary market production center can be expressed as:
\begin{equation}
    \begin{split}
        \Pi_{2 }=& \max\limits_{s \leq x,s \leq q_{2 },q_{2 }\geq 0}  \{ E_{D_{2 } }\left [p_{2 }\min(D_{2 },q_{2 })-g_{2 }(D_{2 }-q_{2 })^{+}+v_{2 }(q_{2 }-D_{2 })^{+}\right]\frac{\theta}{\theta + \Delta  \theta}\\
&- c_{2 }\frac{\theta}{\theta + \Delta  \theta}(q_{2 }-s)-sC+ v_{1 }(x-s)    \},
    \end{split}
\end{equation}

The first term on the right-hand side of the profit function is the expected sales revenue, the second and third terms respectively represent the expected shortage cost and the expected salvage value, the fourth term accounts for the production cost, the fifth term represents the transportation cost from the primary market to the secondary market, the last residual value term explains the value of products retained in the primary market that are not shipped to the secondary market. In addition, the first four terms are calculated in the currency of the primary market.  
 
We rearrange the profit function in equation (1) to obtain:
\begin{equation}
    \begin{split}
        \Pi_{2 }=& \max\limits_{q_{2 }\geq 0}  \{E_{D_{2 } }\left[ p_{2 }\min(D_{2 },q_{2 })-g_{2 }(D_{2 }-q_{2 })^{+}+v_{2 }(q_{2 }-D_{2 })^{+}-c_{2 }q_{2 }\right ] \frac{\theta}{\theta + \Delta  \theta}\\
&+ \max\limits_{s \leq x, s \leq q_{2 }} (c_{2 }\frac{\theta}{\theta + \Delta  \theta}-C-v_{1 })s    \}+v_{1 }x.
    \end{split}
\end{equation}
   
Let $L(x,q_{2 })=\max\limits_{s \leq x, s \leq q_{2 }}\left(c_{2 }\frac{\theta}{\theta + \Delta  \theta}-C-v_{1 }  \right )s  $. It can be seen that if it is cheaper for the secondary market production center to directly produce products than to transfer (i.e., $c_{2 }\frac{\theta}{\theta + \Delta  \theta}< C+v_{1 }$), enterprises will have no motivation to ship anything from the primary market to the secondary market (i.e., $s^{*}=0$). Otherwise, when $c_{2 }\frac{\theta}{\theta + \Delta  \theta}\geq C+v_{1 }$, then, the optimal transshipment strategy of the enterprise is $s^{*}=\min (x,q_{2 })$. Thus, we can obtain the equivalent expression of $L(x,q_{2 })$ as 
\begin{equation*} 
L(x,q_{2 })=\left (c_{2 }\frac{\theta}{\theta + \Delta  \theta}-C-v_{1 }\right )^{+}\min (x,q_{2 }).
\end{equation*}

Scarf (1958) first solved the newsvendor problem that only knows the mean and variance information of the demand by using the method of distributionally robust optimization. In fact, from the profit function in (2) we can observe that if $c_{2 }\frac{\theta}{\theta + \Delta  \theta}< C+v_{1 }$, $q_{2 }^{*}$ corresponds to the optimal solution of the traditional robust newsvendor problem (see Scarf 1958) for the secondary market, which is:
\begin{eqnarray} 
\hat{q_{2 }^{*}}=
\begin{cases}
E(D_{2 })+\frac{\sigma (D_{2 })}{2}\left( \sqrt{\frac{p_{2 }+g_{2 }-c_{2 }}{c_{2 }-v_{2 }}} - \sqrt{\frac{c_{2 }-v_{2 }}{p_{2 }+g_{2 }-c_{2 }} }   \right),& \frac{\sigma (D_{2 })}{E(D_{2 })}<\sqrt{\frac{p_{2 }+g_{2 }-c_{2 }}{c_{2 }-v_{2 }}},\\
0,& \frac{\sigma (D_{2 })}{E(D_{2 })}\geq \sqrt{\frac{p_{2 }+g_{2 }-c_{2 }}{c_{2 }-v_{2 }}},\\
\end{cases}
\end{eqnarray}
where $E(D_{2 })$ and $\sigma (D_{2 })$ represent the expectation and variance of the random variable $D_{2 }$, respectively.

The formula (3) explains well how the effective matching between supply and demand affects the choice of an enterprise's sales model. In the optimal sales rule given by (3), the expectation $E (D_{2 })$ and variance $\sigma(D_{2 })$ reflect market demand, $c_{2 }$ reflects supply, and $p_{2 }$ and $g_{2 }$ essentially constitute a control factor between supply and demand. Therefore, the optimal solution obtained from our robust perspective reflects both supply and demand, as well as the prices $p_{2 }$ and $g_{2 }$ for building a bridge between them. We often talk about the relationship between supply and demand, but what kind of relationship is between supply and demand, and how it affects our final decision-making, etc. These questions are not answered by the classic newsvendor problem with a known specific distribution. On the contrary, robust newsvendor provides a very intuitive, explicit and highly explanatory condition, which mainly explains how the relationship between supply and demand affects our choice of optimal strategy. Therefore, if the robust newsvendor method combines the approximate optimality of its decision-making, it is basically more practical than the classic newsvendor.
   
In addition, the optimal inventory given by (3) consists of two terms when $\frac{\sigma (D_{2 })}{E(D_{2 })}<\sqrt{\frac{p_{2 }+g_{2 }-c_{2 }}{c_{2 }-v_{2 }}}$. The first term represents the expected value of demand in the secondary market, and the second term denotes a safety stock to hedge the uncertainty of market demand. We also notice that $\hat{q_{2 }^{*}}$ is independent of exchange rate factors (i.e., the specific value of $\theta$ and $\Delta \theta$) and of the remaining inventory at the primary market, $x$. In this case, we have $s^{*}=0$, and the production strategy of the secondary market production center is $q_{2 }^{*}-s^{*}=\hat{q_{2 }^{*}}$.

\begin{theorem}\label{thm1}
For the secondary market, when $c_{2 }\frac{\theta}{\theta + \Delta  \theta}< C+v_{1 }$, we have $s^{*}=0$ and  there are some thresholds $\ddot{c}_{2 }=\frac{(p_{2 }+g_{2 })E(D_{2 })^{2}+v_{2 }\sigma (D_{2 })^{2} }{\sigma (D_{2 })^{2}+E(D_{2 })^{2}  }  $, 
$\ddot{p}_{2 }=\frac{(c_{2 }-v_{2 })\sigma (D_{2 })^{2}}{E(D_{2 })^{2}}+c_{2 }-g_{2 }    $, 
$\ddot{g}_{2 }=\frac{(c_{2 }-v_{2 })\sigma (D_{2 })^{2}}{E(D_{2 })^{2}}+c_{2 }-p_{2 }    $, 
$\ddot{v}_{2 }=c_{2 }-\frac{(p_{2 }+g_{2 }-c_{2 })E(D_{2 })^{2}}{\sigma (D_{2 })^{2}}   $, 
$\ddot{\sigma} (D_{2 })=\sqrt{\frac{p_{2 }+g_{2 }-c_{2 }}{c_{2 }-v_{2 }}} E(D_{2 })  $,
 and $\ddot{E}(D_{2 })=\sqrt{\frac{c_{2 }-v_{2 }}{p_{2 }+g_{2 }-c_{2 }}} \sigma (D_{2 })    $.
 If and only if $c_{2 }<\ddot{c}_{2 }$, $p_{2 }>\ddot{p}_{2 }$, $g_{2 }>\ddot{g}_{2 }$, $v_{2 }>\ddot{v}_{2 }$, $\sigma (D_{2 })<\ddot{\sigma} (D_{2 })$, or $E(D_{2 })>\ddot{E}(D_{2 })$, then the enterprise should carry out production in the production center.

The above six thresholds $\ddot{c}_{2 }$, $\ddot{p}_{2 }$, $\ddot{g}_{2 }$, $\ddot{v}_{2}$, $\ddot{\sigma} (D_{2 })$, and $\ddot{E}(D_{2 })$  are equivalent, and the optimal inventory $q_{2 }^{*}$ is the optimal production quantity $q_{2 }^{*}-s^{*}$, which are all $E(D_{2 })+\frac{\sigma (D_{2 })}{2}\left( \sqrt{\frac{p_{2 }+g_{2 }-c_{2 }}{c_{2 }-v_{2 }}} - \sqrt{\frac{c_{2 }-v_{2 }}{p_{2 }+g_{2 }-c_{2 }} }   \right)$.
\end{theorem}

Theorem~{\upshape\ref{thm1}} indicates that whether an enterprise invests in production in a secondary market production centre depends on a trade-off between several factors. Conversely, if $c_{2 }\frac{\theta}{\theta + \Delta  \theta}\geq C+v_{1 }$, it means that it is cheaper to ship products from the primary market rather than to produce them in the secondary market. The transshipment strategy of the enterprise is $s^{*}=\min (x,q_{2 })$. At the same time, $x>q_{2 }$ demonstrates that there are too many products available for transshipment in the primary market, resulting in the planned inventory that can only be transshipped in the secondary market, i.e., $s^{*}=q_{2 }$. Therefore, the corresponding robust optimal solution $q_{2 }^{*}$ is obtained as follows:
\begin{eqnarray}
\tilde{q_{2 }^{*}}=
\begin{cases}
E(D_{2 })+\frac{\sigma (D_{2 })}{2}
\left( \sqrt{K} 
-\sqrt{\frac{1}{K}}   \right),
& \frac{\sigma (D_{2 })}{E(D_{2 })}
<\sqrt{K},\\
0,& \frac{\sigma (D_{2 })}{E(D_{2 })}\geq \sqrt{K},\\
\end{cases}
\end{eqnarray}
where $
K=\frac{(p_{2 }+g_{2 })\frac{\theta}{\theta +\Delta \theta}-C-v_{1 }}{C+v_{1 }-v_{2 }\frac{\theta}{\theta +\Delta \theta}}$.

Moreover, in the presence of adequate demand in the secondary market, the optimal transshipment strategy is all the remaining inventory after the end of the sales season in the primary market, i.e., $s^{*}=x$. After calculation, the optimal inventory $q_{2 }^{*}$ is the same as in the case of $s^{*}=0$, which is $\hat{q_{2 }^{*}}$.

By comparing (3) and (4), we can conclude that $\hat{q_{2 }^{*}} \leq \tilde{q_{2 }^{*}}$. Hence, the optimal inventory at the secondary market is to reserve to $q_{2 }^{*}=\max \left(\hat{q_{2 }^{*}},\min \left(x,\tilde{q_{2 }^{*}}\right)\right)$, and the optimal transshipment strategy from the primary market is given by $s^{*}=\min (x,q_{2 }^{*})$. As a supplement to the research of Kouvelis and Gutierrez [23], this paper provides enterprises with an operation strategy under a two-market stochastic system without knowing the demand distribution.

\begin{theorem}\label{thm2}

(i) If $c_{2 }\frac{\theta}{\theta + \Delta  \theta}< C+v_{1 }$, we have $q_{2 }^{*}=\hat{q_{2 }^{*}}$, $s^{*}=0$, and $q_{2 }^{*}-s^{*}=\hat{q_{2 }^{*}}$.

\noindent
(ii) If $c_{2 }\frac{\theta}{\theta + \Delta  \theta}\geq C+v_{1 }$, 
 
(a) for $x<\hat{q_{2 }^{*}}$, we have $q_{2 }^{*}=\hat{q_{2 }^{*}}$, $s^{*}=x$, and $q_{2 }^{*}-s^{*}=\hat{q_{2 }^{*}}-x$;

(b) for $\hat{q_{2 }^{*}}\leq x< \tilde{q_{2 }^{*}}$, we have $q_{2 }^{*}=x$, $s^{*}=x$, and $q_{2 }^{*}-s^{*}=0$;

(c) for $x \geq \tilde{q_{2 }^{*}}$, we have $q_{2 }^{*}=\tilde{q_{2 }^{*}}$, $s^{*}=\tilde{q_{2 }^{*}}$, and $q_{2 }^{*}-s^{*}=0$.
\end{theorem}

Theorem~{\upshape\ref{thm2}} shows that the threshold conditions $c_{2 }\frac{\theta}{\theta + \Delta  \theta}< C+v_{1 }$ and $c_{2 }\frac{\theta}{\theta + \Delta  \theta}\geq C+v_{1 }$ for enterprises first when deciding whether to implement transshipment, do not take the production cost $c_{1 }$ of the primary market into consideration(unrelated to $c_{1 }$, since the product is already produced). We only need to consider the unit cost $C$ of the product in transit and the unit residual value $v_{1 }$ of the remaining product. These two factors are combined with secondary market production cost $c_{2 }$ and exchange rate to determine the early transshipment strategy. The larger $c_{2 }$, the less economical it is to produce in the secondary market. Therefore, in order to reduce regenerative output, enterprises should choose transshipment. However, the higher the transshipment costs $C$ or residual value $v_{1 }$, the easier it is to meet the condition $c_{2 }\frac{\theta}{\theta + \Delta  \theta}< C+v_{1 }$, and it will be more beneficial for the enterprise to keep the products available for transportation in the primary market.

\begin{figure}[h]%
\centering
\includegraphics[width=0.99\textwidth]{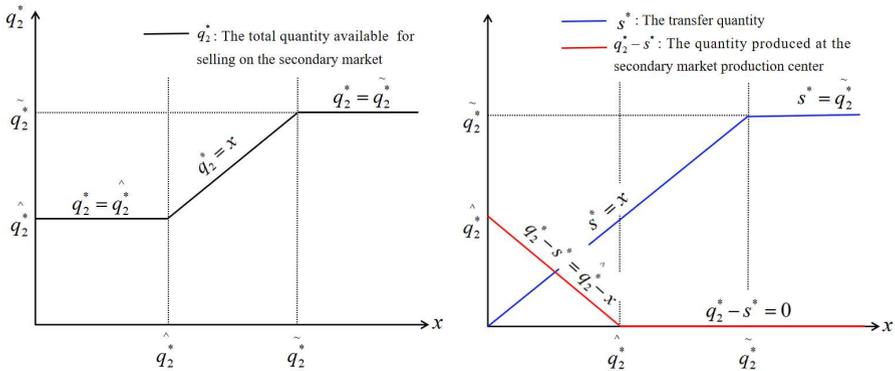}
\caption{Optimal operating strategy on the secondary market}\label{fig3}
\end{figure}

Figure~\ref{fig3} explains the impact of surplus inventory $x$ on the operation strategy ($q_{2 }^{*}$, $q_{2 }^{*}-s^{*}$, and $s^{*}$) of the secondary market. In the case that there is no product available for export in the primary market (i.e., $x=0$), transshipment does not occur (i.e., $s^{*}=0$). In order to meet the demand of the secondary market, enterprises will produce all the required quantities after balancing the related risks. The solution of the robust newsvendor problem provides an optimal inventory $q_{2 }^{*}=\hat{q_{2 }^{*}}$ for the secondary market. Specifically, if and only if the relationship $\frac{\sigma (D_{2 })}{E(D_{2 })}<\sqrt{\frac{p_{2 }+g_{2 }-c_{2 }}{c_{2 }-v_{2 }}}$ between supply and demand is satisfied, the enterprise will make the decision to reproduce, otherwise, it will choose not to produce due to factors such as high cost or excessive demand risk (see Theorem~{\upshape\ref{thm1}}). 
  
The remaining inventory in the primary market per unit can be used to satisfy the demand in the secondary market with a unit production cost of $\frac{\theta + \Delta  \theta}{\theta }(C+v_{1 })$ (in secondary market currency units). The enterprise considers the characteristics of the demand and cost of the secondary market to make a corresponding optimal strategy $q_{2 }^{*}=\tilde{q_{2 }^{*}}$. Any remaining inventory $x$ below $\tilde{q_{2 }^{*}}$ should be transferred. On the one hand, if $x<\hat{q_{2 }^{*}}$, the production center must produce the required additional replenishment quantity in order to make the available inventory in the secondary market reach $\hat{q_{2 }^{*}}$ at the beginning of the selling season. On the other hand, if $x>\tilde{q_{2 }^{*}}$, only the quantity $\tilde{q_{2 }^{*}}$ expected by the secondary market can be transferred, and the untransferred part (i.e., $x-\tilde{q_{2 }^{*}}$) is treated as a unit residual value $v_{1 }$ in the primary market. In addition, a consistent observation with our intuition is that $s^{*}$ and $q_{2 }^{*}$ are increasing in $x$, while $q_{2 }^{*}-s^{*}$ is decreasing in $x$.

\begin{proposition}
For the optimal inventory in the secondary market, 

(i) $\hat{q_{2 }^{*}}$ and $\tilde{q_{2 }^{*}}$ are increasing in $E(D_{2 })$, $\sigma(D_{2 })$, and $p_{2 }$ (i.e., $\frac{d\hat{q_{2 }^{*}}}{d E(D_{2 })}>0$, $\frac{d\tilde{q_{2 }^{*}}}{d E(D_{2 })}>0$, $\frac{d\hat{q_{2 }^{*}}}{d \sigma(D_{2 })}>0$, $\frac{d\tilde{q_{2 }^{*}}}{d \sigma(D_{2 })}>0$, $\frac{d\hat{q_{2 }^{*}}}{d p_{2 }}>0$, and $\frac{d\tilde{q_{2 }^{*}}}{d p_{2 }}>0$). 

(ii) $\hat{q_{2 }^{*}}$ is independent of $\theta$ and $\Delta \theta$. 

(iii) $\tilde{q_{2 }^{*}}$ is decreasing in $\Delta \theta$, but independent in $\theta$ if $\Delta \theta=0$, increasing in $\theta$ if $\Delta \theta >0$, and decreasing in $\theta$ if $\Delta \theta <0$.
\end{proposition}

Proposition 3 analyzes some of the factors affecting the quantity of inventory in the secondary market. Among them, (i) indicates that $E(D_{2 })$ and $\sigma(D_{2 })$ have a positive effect on the enterprise. Even if the demand fluctuates greatly and the uncertainty risk is high, the enterprise will increase the inventory of the secondary market in order to meet the market demand as far as possible without causing loss of stockout. Simultaneously, a very intuitive conclusion is that the increase in sales price $p_{2 }$ will also tempt enterprises to increase inventory. (ii) and (iii) show the different effects of exchange rate factors on optimal inventory in the secondary market under different circumstances.

\begin{proposition}
The worst-case profit of the secondary market production center is:

\noindent
(i) If $c_{2 }\frac{\theta}{\theta + \Delta  \theta}< C+v_{1 }$, we have $\Pi_{2 }^{*}=\left (W_{1 }-c_{2 }\hat{q_{2 }^{*}}    \right )\frac{\theta}{\theta + \Delta \theta}+ v_{1 }x $.

\noindent
(ii) If $c_{2 }\frac{\theta}{\theta + \Delta  \theta}\geq C+v_{1 }$, 

(a) for $x<\hat{q_{2 }^{*}}$, we have $\Pi_{2 }^{*}=\left (W_{1 }-c_{2 }\hat{q_{2 }^{*}}    \right )\frac{\theta}{\theta + \Delta \theta}+ c_{2 }\frac{\theta}{\theta + \Delta \theta}x -Cx$;

(b) for $\hat{q_{2 }^{*}}\leq x< \tilde{q_{2 }^{*}}$, we have $\Pi_{2 }^{*}=W_{2 }\frac{\theta}{\theta + \Delta \theta}-Cx $;

(c) for $x \geq \tilde{q_{2 }^{*}}$, we have $\Pi_{2 }^{*}=W_{3 }\frac{\theta}{\theta + \Delta \theta}-(C+v_{1 })\tilde{q_{2 }^{*}}+ v_{1 }x $.

And where 
\[
W_{1 }=p_{2 }\min \left (D_{2 },\hat{q_{2 }^{*}}  \right )-g_{2 }\left (D_{2 }-\hat{q_{2 }^{*}}  \right )^{+}+ v_{2 }    \left (\hat{q_{2 }^{*}} -D_{2 } \right )^{+},
\]
\[
W_{2 }=p_{2 }\min \left (D_{2 },x  \right )-g_{2 }\left (D_{2 }-x  \right )^{+}+ v_{2 }    \left (x -D_{2 } \right )^{+},
\] 
\[
W_{3 }=p_{2 }\min \left (D_{2 },\tilde{q_{2 }^{*}}  \right )-g_{2 }\left (D_{2 }-\tilde{q_{2 }^{*}}  \right )^{+}+ v_{2 }    \left (\tilde{q_{2 }^{*}} -D_{2 } \right )^{+}.
\]
   
\end{proposition}

Note that although the optimal inventory $\hat{q_{2 }^{*}}$ and $\tilde{q_{2 }^{*}} $ in the secondary market are not related to $x$, the worst-case profit $\Pi_{2 }^{*}$ is affected by it and increasing in $x$ (i.e., $\partial{\Pi_{2 }^{*}}/\partial{x}>0$).

\subsection{Optimal production strategy in the primary market}\label{subsec3}
We now consider the production strategy for the primary market, and let $q_{1 }$ be the quantity to be produced. Then, the profit function $\Pi_{1 }$ for the enterprise can be written as:
\begin{equation}
    \begin{split}
        \Pi_{1 }=\max\limits_{q_{1 }\geq 0}\left \{ E_{D_{1 }} \left [p_{1 }\min(D_{1 },q_{1 })-g_{1 }(D_{1 }-q_{1 })^{+}     -c_{1 }q_{1 }+\Pi_{2 }\left ( q_{1 }-D_{1 }   \right )^{+}  \right]  \right \},
    \end{split}
\end{equation}

The first term in the above profit function represents the sales profit, the second term accounts for the shortage cost, the third term represents the production cost, and the last term is the expected total profit of the secondary market.

Same as section~\ref{subsec2}, if $c_{2 }\frac{\theta}{\theta + \Delta  \theta}< C+v_{1 }$, the enterprise has no incentive to transfer any products (i.e., $s^{*}=0$), and the optimal robust production strategy for the primary market is:
\begin{eqnarray}
\hat{q_{1 }^{*}}=
\begin{cases}
E(D_{1 })+\frac{\sigma (D_{1 })}{2}\left( \sqrt{\frac{p_{1 }+g_{1 }-c_{1 }}{c_{1 }-v_{1 }}} - \sqrt{\frac{c_{1 }-v_{1 }}{p_{1 }+g_{1 }-c_{1 }} }   \right),& \frac{\sigma (D_{1 })}{E(D_{1 })}<\sqrt{\frac{p_{1 }+g_{1 }-c_{1 }}{c_{1 }-v_{1 }}},\\
0,& \frac{\sigma (D_{1 })}{E(D_{1 })}\geq \sqrt{\frac{p_{1 }+g_{1 }-c_{1 }}{c_{1 }-v_{1 }}}.\\
\end{cases}
\end{eqnarray}
where $E(D_{1 })$ and $\sigma (D_{1 })$ represent the expectation and variance of the random variable $D_{1 }$, respectively.

If $c_{2 }\frac{\theta}{\theta + \Delta  \theta}\geq C+v_{1 }$, and there is sufficient demand in the secondary market, the enterprise will transfer all surplus inventory in the primary market (i.e., $s^{*}=(q_{1 }-D_{1 })^{+}$). In this case, the optimal robust production strategy for the primary market is:
\begin{eqnarray}
\tilde{q_{1 }^{*}}=   
\begin{cases}
E(D_{1 })+\frac{\sigma (D_{1 })}{2}\left( \sqrt{\overline{K}} - \sqrt{\frac{1}{\overline{K}}}   \right),& \frac{\sigma (D_{1 })}{E(D_{1 })}<\sqrt{\overline{K}},\\
0,& \frac{\sigma (D_{1 })}{E(D_{1 })}\geq \sqrt{\overline{K}},\\
\end{cases}
\end{eqnarray}
where $\overline{K}=\frac{p_{1 }+g_{1 }-c_{1 }}{c_{1 }+C-c_{2 }\frac{\theta}{\theta+ \Delta \theta} }$.

It is not difficult to find that when the transshipment and the transshipment quantity cannot meet the expected demand, the production strategy $\tilde{q_{1 }^{*}}$ of the primary market is different from the situation where $\hat{q_{1 }^{*}}$ is independent of the relevant parameters of the secondary market without transshipment. Conversely, $\tilde{q_{1 }^{*}}$ is related to production costs $c_{2 }$ and exchange rates ($\theta$ and $\Delta \theta$). 

If $s^{*}=q_{2 }$, the secondary market production center does not need to produce, and the optimal inventory in the secondary market is the same as $\hat{q_{1 }^{*}}$.

Actually, in the absence of a secondary market, the profit function of the primary market production center is as follows:
\begin{equation*}
    \begin{split}
        \Pi_{1 }^{0}=p_{1 }\min(D_{1 },q_{1 })-g_{1 }(D_{1 }-q_{1 })^{+}+ v_{1 }(q_{1 }-D_{1 })^{+}    -c_{1 }q_{1 },
    \end{split}
\end{equation*}
and the optimal production strategy in this case is $q_{1 }^{*}=\hat{q_{1 }^{*}}$, which corresponds to the situation that the secondary market exists but no transshipment. In addition, according to condition $c_{2 }\frac{\theta}{\theta + \Delta  \theta}\geq C+v_{1 }$, we can observe that the relationship $\tilde{q_{1 }^{*}} \geq \hat{q_{1 }^{*}}$ is satisfied (see Figure~\ref{fig4}), and the result is verbally stated in the following theorem:

\begin{figure}[h]%
\centering
\includegraphics[width=0.62\textwidth]{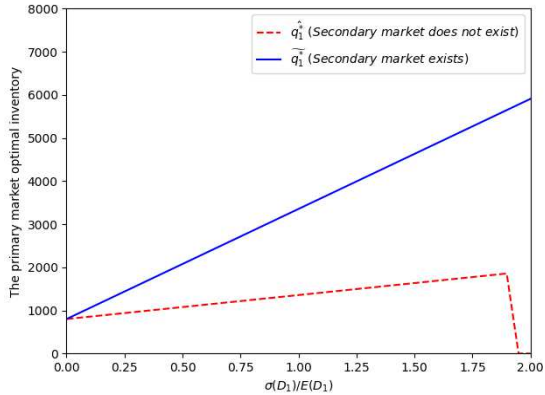}
\caption{Comparison of optimal inventory in primary market with or without secondary market (Notes. Parameter values are $p_{1 }=10$, $g_{1 }=6$, $v_{1 }=2$, $c_{1 }=5$, $c_{2 }=6$, $C=1$, $E(D_{1 })=800$, $\theta =6.70$, and $\Delta \theta=0.30$)}\label{fig4}
\end{figure}

\begin{theorem}\label{thm5}
The existence of a secondary market, regardless of whether there is a production center in that market, will result in enterprises having more motivation to produce more (in a weak sense, there is an equal possibility) in the primary market. That is, the quantity produced by the primary market production center will be greater than the quantity produced in the absence of the secondary market.
\end{theorem}

What we want to point out is that the conclusion in Theorem~{\upshape\ref{thm5}} is consistent with the conclusion given by Kouvelis and Gutierrez (1997). This is actually in line with people’s common sense. After the enterprise balances supply and demand risks, the remaining inventory on the primary market may be sold on the secondary market, which indirectly reduces the economic risk of oversupply of products on the primary market. In this case, therefore, greater product coverage would give enterprises an incentive to produce more in the primary market.

\begin{proposition}  
For the optimal inventory in the primary market, 

(i) $\hat{q_{1 }^{*}}$ and $\tilde{q_{1 }^{*}}$ are increasing in $E(D_{1 })$, $\sigma (D_{1 })$, and $p_{1 }$ (i.e., $\frac{d\hat{q_{1 }^{*}}}{d E(D_{1 })}>0$, $\frac{d\tilde{q_{1 }^{*}}}{d E(D_{1 })}>0$, $\frac{d\hat{q_{1 }^{*}}}{d \sigma(D_{1 })}>0$, $\frac{d\tilde{q_{1 }^{*}}}{d \sigma(D_{1 })}>0$, $\frac{d\hat{q_{1 }^{*}}}{d p_{1 }}>0$, and $\frac{d\tilde{q_{1 }^{*}}}{d p_{1 }}>0$).  

(ii) $\hat{q_{1 }^{*}}$ is independent of $\theta$ and $\Delta \theta$. 

(iii) $\tilde{q_{1 }^{*}}$ is decreasing in $\Delta \theta$, but independent in $\theta$ if $\Delta \theta=0$, increasing in $\theta$ if $\Delta \theta >0$, and decreasing in $\theta$ if $\Delta \theta <0$.

\end{proposition}

\begin{figure}[h]%
\centering
\includegraphics[width=0.99\textwidth]{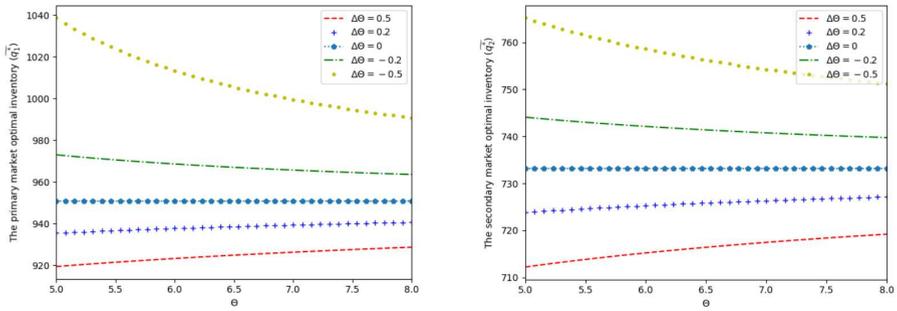}
\caption{The effect of exchange rate on $\tilde{q_{1 }^{*}}$ and $\tilde{q_{2 }^{*}}$ (Notes. Parameter values are $p_{1 }=p_{2 }=10$, $g_{1 }=g_{2 }=6$, $v_{1 }=v_{2 }=2$, $c_{1 }=5$, $c_{2 }=6$, $C=1$, $E(D_{1 })=800$, $E(D_{2 })=600$, $\theta =6.70$, and $\Delta \theta=0.30$)}\label{fig5}
\end{figure}

Similar to Proposition 3, Proposition 6 analyses some of the factors affecting primary market production strategy. In addition, Fig.~\ref{fig5} explains the influence of exchange rate on optimal inventory $\tilde{q_{1 }^{*}}$ and $\tilde{q_{2 }^{*}}$, respectively. We should note that Fig.~\ref{fig5}a and \ref{fig5}b have the same trend. As shown in Fig.~\ref{fig5}a, with the increase of the risk of exchange rate fluctuations (i.e., $\lvert  \Delta \theta \rvert $ $\uparrow$), the enterprise will reduce the optimal production quantity (i.e., $\tilde{q_{1 }^{*}}(\lvert \Delta \theta \rvert)\downarrow$) for the primary market in order to avoid the loss of a large amount of surplus inventory. If $\Delta \theta=0$, that is, there is no risk of exchange rate fluctuations, then the optimal production strategy in the primary market is not affected by exchange rate factors. If $\Delta \theta >0$, the currency in the secondary market appreciates, $c_{2 }\frac{\theta}{\theta + \Delta  \theta}$ is increasing in $\theta$, and condition $c_{2 }\frac{\theta}{\theta + \Delta  \theta}\geq C+v_{1 }$ is easier to get. At this time, the production cost $c_{2 }\frac{\theta}{\theta + \Delta  \theta}$ in the secondary market is higher than the cost $C+v_{1 }$ of transferring inventory in the primary market, which stimulates enterprises to increase production in the primary market (i.e., $\Delta \theta >0 \rightarrow \tilde{q_{1 }^{*}}(\theta)\uparrow$). However, if $\Delta \theta <0$, the currency in the secondary market depreciates, $c_{2 }\frac{\theta}{\theta + \Delta  \theta}$ is decreasing in $\theta$, and condition $c_{2 }\frac{\theta}{\theta + \Delta  \theta}\geq C+v_{1 }$ is more difficult to satisfy, that is, the enterprise will be more inclined to not transfer inventory in the primary market. Thus, in order to match supply and demand well, enterprises will reduce the production quantity of the primary market (i.e., $\Delta \theta <0 \rightarrow \tilde{q_{1 }^{*}}(\theta)\downarrow$).

\subsection{Robustness of optimal production strategies in two markets}\label{subsec4}
According to the robust ordering rules proposed by Scarf (1958), decision makers or managers will choose not to order or not to sell (the quantity is zero) under certain conditions. This ordering rule has been pointed out as too conservative. However, Perakis and Roels (2008) and Fu et al. (2018) show that the robust order quantity is quite close to the risk-neutral order quantity under the known distribution. For the global robust newsvendor problem where both market demands are random, we also conducted robustness analysis.
   
Based on Andersson et al. (2013), we sampled the demand distribution of the two markets as shown below. We first consider the sampling of random demand $D_{2 }$ in the secondary market: randomly select $n=10$ different ranking values in interval $[0,3q_{2 }]$ (the upper limit $3q_{2 }$ here is not unique), and provide 10 values $d_{2 }^{1}, d_{2 }^{2}, \cdots , d_{2 }^{10},$ as the random demand $D_{2 }$. Then extract 10 new values $s_{2 }^{1}, s_{2 }^{2}, \cdots , s_{2 }^{10},$ on the interval [0,1] and regularize them to $\rho_{2 }^{i}=\frac{s_{2 }^{i}}{\sum\nolimits_{j=1 }^{10}s_{2 }^{j}}$. Therefore, we get the sampling distribution of $D_{2 }$ with value $d_{2 }^{1}, d_{2 }^{2},\cdots , d_{2 }^{10},$ and probability $\rho_{2 }^{1}, \rho_{2 }^{2},\cdots , \rho_{2 }^{10}$. Then, we obtained the mean $E(D_{2 })=\sum\nolimits_{i=1 }^{10}\rho_{2 }^{i}d_{2 }^{i}$ and standard variance $\sigma (D_{2 })=\sqrt{\sum\nolimits_{i=1 }^{10}\rho_{2 }^{i}(d_{2 }^{i}-E(D_{2 }))}$ through simple algebraic operations. The consideration of sampling distribution of random demand $D_{1 }$ is the same as that of $D_{2 }$, which will not be repeated here.

\begin{figure}[h]%
\centering
\includegraphics[width=0.99\textwidth]{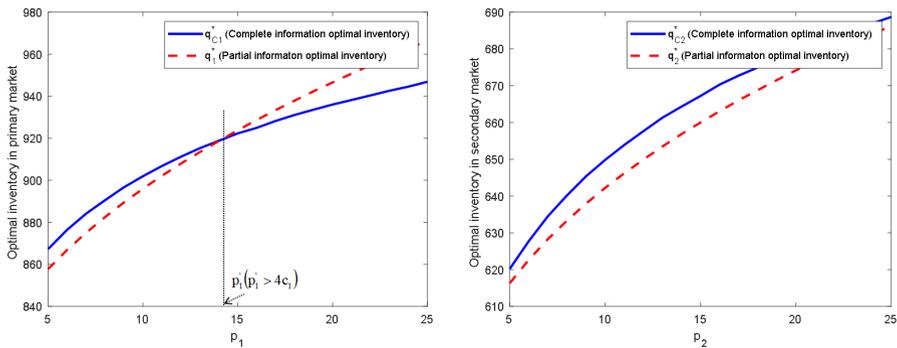}
\caption{Comparison between the complete and partial information (robust) optimal inventory}\label{fig6}
\end{figure}

To evaluate the performance of the global robust newsvendor problem, we consider the following problem setting: $p_{1 }=p_{2 }=15$, $c_{1 }=c_{2 }=5$, $g_{1 }=g_{2 }=6$, $v_{1 }=v_{2 }=3$, $C=2$, $E(D_{1 })=800$, $\sigma(D_{1 })=100$, $E(D_{2 })=600$, $\sigma(D_{2 })=80$, $\theta =6.70$, and $\Delta \theta =0.30$. We use the above steps to sample 10,000 distribution samples, and by changing the sales price $p_{1 }$ and $p_{2 }$, we obtain the comparison curve between the robust production strategy $(q_{1 }^{*},q_{2 }^{*})$ and the production strategy $(q_{C1 }^{*},q_{C2 }^{*})$ under the bivariate normal distribution as shown in Figure~\ref{fig6}.

From the comparison in Figure~\ref{fig6}, we can see that the robust production strategies of the two markets are very close to the production strategies under complete information, which is consistent with the results of Perakis and Roels (2008) and Fu et al. (2018). Interestingly, we find that robust production strategy in the primary market is not always lower or higher than complete information (see Figure~\ref{fig6}a). When $p_{1 }>p_{1 }^{'}$ (where $p_{1 }^{'}>2c_{1 }$), $q_{1 }^{*}$ is more than $q_{C1 }^{*}$, otherwise the opposite. This indicates that if $p_{1 }$ is high relative to $c_{1 }$, the enterprise is willing to plan more inventory in the primary market to ensure the demand on the primary market first. Because even if the planned quantity is too large, resulting in too much surplus inventory in the primary market, but with the buffering effect of the secondary market, the enterprise will be more profitable. This result is different from the previous one-stage robust newsvendor decision-making problem, which is always more conservative (lower number under robust decision-making) than the decision with complete information. Therefore, in the sense of the research question in this paper, robustness is not a conservative in nature, but an environmental adaptability and a path choice.

\section{Two-market single product global robust newsvendor problem under demand-related}\label{sec5}
In our discussion in section~\ref{sec3}, the demands between the two markets are assumed to be totally independent. However, these two markets are linked by the same product, and it seems that $D_{1 }$ and $D_{2 }$ should be related at least to some degree. For example, changes in the market environment, corporate brand image or consumer preferences, all have an impact on both sales. In view of this situation, this section relaxes the assumption of demand independence. Meanwhile, the enterprise's goals and all other settings in the model remain unchanged.

Under demand-related settings, considering the net profit $\Pi$ of the enterprise, denoted by
\begin{equation}
    \begin{split}
        \Pi=& \max\limits_{s^{C}\leq (q_{1 }^{C}-D_{1 })^{+}; s^{C}\leq q_{2 }^{C};q_{1 }^{C},q_{2 }^{C}\geq 0} E_{D_{1 },D_{2} } \left [P-G+V-T       \right],
    \end{split}
\end{equation}
where
\[
P=P(D_{1 },q_{1 }^{C},D_{2 },q_{2 }^{C},\theta , \Delta \theta)=P_{1 }\min (D_{1 },q_{1 }^{C})+P_{2 }\frac{\theta}{\theta + \Delta  \theta}\min (D_{2 },q_{2 }^{C}),
\]
\[
G=G(D_{1 },q_{1 }^{C},D_{2 },q_{2 }^{C},\theta , \Delta \theta)=g_{1 }(D_{1 }-q_{1 }^{C})^{+}+g_{2 }\frac{\theta}{\theta + \Delta  \theta}(D_{2 }-q_{2 }^{C})^{+},
\]
\[
V=V(D_{1 },q_{1 }^{C},D_{2 },q_{2 }^{C},\theta , \Delta \theta ,s^{C})=v_{1 }\left [(q_{1 }^{C}-D_{1 })^{+}-s \right ]+v_{2 }\frac{\theta}{\theta + \Delta  \theta}(q_{2 }^{C}-D_{2 })^{+},
\]
\[
T=T(q_{1 }^{C},q_{2 }^{C},\theta , \Delta \theta ,s^{C})=c_{1 }q_{1 }^{C}+c_{2 }\frac{\theta}{\theta + \Delta  \theta}(q_{2 }^{C}-s^{C})+s^{C}C.
\]

The net profit function $\Pi$ in the above expression consists of four terms. The first term indicates the sales profit. The second term represents the loss of stock when supply exceeds demand. The third term explains the residual value when supply exceeds demand, and the fourth term is the sum of production cost and transshipment cost. The above four terms are all calculated in the currency of the primary market.
 
We are interested in the situation where distributions of demand between the two markets are not known precisely when designing the transnational operation strategy. Instead, we only have information about the means, variances, and covariance of random demand $D_{1 }$ and $D_{2 }$, which are given by a moments matrix $\Sigma$:
\[
\Sigma=\left(
 \begin{array}{ccc}
   E(D_{1 }^{2}) & E(D_{1 }D_{2 }) & E(D_{1 })\\
   E(D_{1 }D_{2 }) & E(D_{2 }^{2}) & E(D_{2 })\\
   E(D_{1 }) & E(D_{2 }) & 1\\
 \end{array}
\right).               
\]
   
Based on Fu et al. (2018) using cross moments to capture the dependence of price and demand in a random market, we here use $E(D_{1 }D_{2 })$ in the moments matrix $ \Sigma$ to represent the correlation between random demand in the two markets. According to the nonnegativity of demand (i.e., $D_{1 }, D_{2 }\geq 0$), it is easy to know that $\Sigma$ is a positive semidefinite matrix. Let 
\[
\Phi(\Sigma_{0 })=\left \{(D_{1 },D_{2 })\lvert D_{1 },D_{2 } \geq 0, D_{1 },D_{2 } \ have\ moments\ matrix\ \Sigma_{0 }=\Sigma \right \}
\]
denote the distribution space with the prescribed moments conditions, where $ \Sigma_{0 }$ is given. Therefore, we evaluate the strategy that we need to against the worst possible distribution in $\Phi(\Sigma_{0 })$.  In other words, the value information of the first and second moments $E(D_{1 })$, $E(D_{2 })$, $E(D_{1 }^{2})$, $E(D_{2 }^{2})$, $E(D_{1 }D_{2 }),$ of the required should be included in $ \Sigma_{0 }$.

The problem faced by the enterprise is to maximize the worst-case profit with the moment constraint $\Phi(\Sigma_{0 })$, by choosing the optimal production strategy $(q_{1 }^{C*},q_{2 }^{C*})$ and transshipment strategy $s^{C*}$. Thus, the global robust newsvendor problem can be formulated as:
\begin{equation}
    \begin{split}
        & \max\limits_{q_{1 }^{C},q_{2 }^{C}\geq 0}\{ -c_{1 }q_{1 }^{C}- c_{2 }\frac{\theta}{\theta + \Delta  \theta}q_{2 }^{C}+\max\limits_{s^{C}\leq (q_{1 }^{C}-D_{1 })^{+}; s^{C}\leq q_{2 }^{C}}\left (c_{2 }\frac{\theta}{\theta + \Delta  \theta}-C-v_{1 }\right )s^{C}\\
& +\inf\limits_{(D_{1 },D_{2 })\in \Phi(\Sigma_{0 })}E_{D_{1 },D_{2 }}\left [ H_{1 }(D_{1 },q_{1 }^{C})+H_{2 }(D_{2 },q_{2 }^{C},\theta , \Delta \theta)\right ]          \},
    \end{split}
\end{equation}
where
\[
H_{1 }(D_{1 },q_{1 }^{C})=p_{1 }\min (D_{1 },q_{1 }^{C})-g_{1 }(D_{1 }-q_{1 }^{C})^{+}+v_{1 }(q_{1 }^{C}-D_{1 })^{+},   
\]
\[
H_{2 }(D_{2 },q_{2 }^{C},\theta , \Delta \theta)=\left [ p_{2 }\min (D_{2 },q_{2 }^{C})-g_{2 }(D_{2 }-q_{2 }^{C})^{+}+v_{2 }(q_{2 }^{C}-D_{2 })^{+} \right ]\frac{\theta}{\theta + \Delta  \theta}.
\]

In the newsvendor problem that only knows the mean and variance information of a single demand, Scarf (1958) first solved this problem by using the method of distributionally robust optimization. In recent decades, the application of robust newsvendor has been widely promoted. However, most of the previous robust newsvendor models only focus on the uncertainty of demand or supply, while few models focus on two uncertain factors, such as ``supply + demand", ``price + demand", and ``demand + demand". That is, the global robust newsvendor problem here is more challenging. The distributed robust Stackelberg game model developed by Fu et al. (2018) has a particular impact on the development of our research problem. The author extends Scarf's robust newsvendor formula to a more general setting under the condition that the price and demand are uncertain and only the first and second moments are available, and the retailer's optimal ordering decision is derived. The research focus of this paper is to provide enterprises with a stable operation strategy in a two-market stochastic system.

An obvious conclusion that can be drawn from equation (9) is that if $c_{2 }\frac{\theta}{\theta + \Delta  \theta}<C+v_{1 }$, there is $s^{C*}=0$. Now we analyze the non-trivial situation $c_{2 }\frac{\theta}{\theta + \Delta  \theta}\geq C+v_{1 }$. In this case, enterprises can have the incentive to transfer sales. Using arguments similar to section~\ref{sec3}, the optimal transfer strategy from the primary market is $s^{C*}=\min \left [(q_{1 }^{C}-E(D_{1 }))^{+},q_{2 }^{C}    \right ]$.
 
In summary, if we let    
\[
L^{C}\left((q_{1 }^{C}-D_{1 })^{+},q_{2 }^{C},s^{C}   \right)=\max\limits_{s^{C}\leq (q_{1 }^{C}-D_{1 })^{+}; s^{C}\leq q_{2 }^{C}}\left (c_{2 }\frac{\theta}{\theta + \Delta  \theta}-C-v_{1 }\right )s^{C},
\] 
then the function $L^{C}\left((q_{1 }^{C}-D_{1 })^{+},q_{2 }^{C},s^{C}   \right)$ has an equivalent expression, which is
\[
L^{C}\left((q_{1 }^{C}-D_{1 })^{+},q_{2 }^{C},s^{C}   \right)=\left (c_{2 }\frac{\theta}{\theta + \Delta  \theta}-C-v_{1 }\right )^{+}\min \left ((q_{1 }^{C}-D_{1 })^{+},q_{2 }^{C}   \right )   .
\]

From equation (9), we can observe that if $c_{2 }\frac{\theta}{\theta + \Delta  \theta}<C+v_{1 }$ (i.e., $s^{C*}=0$), the optimal production strategies for the two markets can be obtained. Therefore, Theorem~{\upshape\ref{thm7}} follows.

\begin{theorem}\label{thm7}
Under the demand-related situation of the two-market stochastic system, if $c_{2 }\frac{\theta}{\theta + \Delta  \theta}<C+v_{1 }$ (i.e., $s^{C*}=0$), the optimal production strategy for the primary market is:
\begin{eqnarray*}
\hat{q_{1 }^{C*}}=
\begin{cases}
E(D_{1 })+\frac{\sigma (D_{1 })}{2}\left( \sqrt{\frac{p_{1 }+g_{1 }-c_{1 }}{c_{1 }-v_{1 }}} - \sqrt{\frac{c_{1 }-v_{1 }}{p_{1 }+g_{1 }-c_{1 }} }   \right),& \frac{\sigma (D_{1 })}{E(D_{1 })}<\sqrt{\frac{p_{1 }+g_{1 }-c_{1 }}{c_{1 }-v_{1 }}},\\
0,& \frac{\sigma (D_{1 })}{E(D_{1 })}\geq \sqrt{\frac{p_{1 }+g_{1 }-c_{1 }}{c_{1 }-v_{1 }}},\\
\end{cases}
\end{eqnarray*}
and for the secondary market is:
\begin{eqnarray*}
\hat{q_{2 }^{C*}}=
\begin{cases}
E(D_{2 })+\frac{\sigma (D_{2 })}{2}\left( \sqrt{\frac{p_{2 }+g_{2 }-c_{2 }}{c_{2 }-v_{2 }}} - \sqrt{\frac{c_{2 }-v_{2 }}{p_{2 }+g_{2 }-c_{2 }} }   \right),& \frac{\sigma (D_{2 })}{E(D_{2 })}<\sqrt{\frac{p_{2 }+g_{2 }-c_{2 }}{c_{2 }-v_{2 }}},\\
0,& \frac{\sigma (D_{2 })}{E(D_{2 })}\geq \sqrt{\frac{p_{2 }+g_{2 }-c_{2 }}{c_{2 }-v_{2 }}}.\\
\end{cases}
\end{eqnarray*}
\end{theorem}
(All proofs are in Appendix F.)
 
From the robust equilibrium solution given by Theorem~{\upshape\ref{thm7}}, there is symmetry and independence between $\hat{q_{1 }^{C*}}$ and $\hat{q_{2 }^{C*}}$. They are dependent on the supply and demand parameters of the market, and have nothing to do with another market. In addition, this solution is completely consistent with the equilibrium solutions (3) and (6) when the two demands are statistically independent (i.e., $\hat{q_{1 }^{C*}}=\hat{q_{1 }^{*}},\hat{q_{2 }^{C*}}=\hat{q_{2 }^{*}}$), and can be regarded as the solution of two robust newsvendor problems in a separate market.
  
Conversely, if $c_{2 }\frac{\theta}{\theta + \Delta  \theta}\geq C+v_{1 }$, the transshipment strategy is 
\[
s^{C*}=\min \left [(q_{1 }^{C}-E(D_{1 }))^{+},q_{2 }^{C}    \right ].
\]
   
If we assume $s^{C*}=(q_{1 }^{C}-E(D_{1 }))^{+}$, that is $(q_{1 }^{C}-E(D_{1 }))^{+}\leq q_{2 }^{C}   $, the optimal production strategy in the secondary market is the same as $\widehat{q_{2 }^{C*}}$, and the optimal production strategy in the primary market is the same as $\tilde{q_{1 }^{*}}$ (i.e., $\tilde{q_{1 }^{C*}}=\tilde{q_{1 }^{*}}$). Similarly, if $s^{C*}=q_{2 }^{C}$, we can get the optimal production strategy in the primary market with $\hat{q_{1 }^{C*}}$, and the optimal production strategy in the secondary market with $\tilde{q_{2 }^{*}}$ (i.e., $\tilde{q_{2 }^{C*}}=\tilde{q_{2 }^{*}}$).

According to the previous analysis, when there is a certain correlation between the demands of different markets, the optimal transshipment strategy is $s^{C*}=\min \left ((q_{1 }^{C}-E(D_{1 }))^{+},q_{2 }^{C*}     \right )$, and the optimal production strategy for the secondary market is reserve to $q_{2 }^{C*}=\max \left(\hat{q_{2 }^{C*}},\min \left ((q_{1 }^{C}-E(D_{1 }))^{+},\tilde{q_{2 }^{C*}}     \right ) \right ) $. 

Interestingly, the optimal operational strategy presented in this paper depends only on the first and second moments of two random demands. The correlation term $E(D_{1 }D_{2 })$ does not affect the optimal choice of $\left (\hat{q_{1 }^{C*}},\hat{q_{2 }^{C*}}   \right)$ or $\left (\tilde{q_{1 }^{C*}},\tilde{q_{2 }^{C*}}   \right)$ in the distributionally robust problem. Similar studies that consider demand correlation such as Petruzzi and Dada (2001), but they do not directly explain its impact on optimal decision-making. It is also worth noting that the robust newsvendor problem in Scarf (1958) is a special case of our global robust newsvendor problem. In this case, the enterprise only produces and sells in one market. Therefore, when there are two demands that are random, our results directly extend Scarf's (1958) robust newsvendor problem to a two-dimensional setting.

\begin{theorem}
The correlation between markets has no effect on the production and transshipment strategies of the enterprise, i.e.,  $q_{1 }^{C*}=q_{1 }^{*}$, $q_{2 }^{C*}=q_{2 }^{*}$, and $s^{C*}=s^{*}$.
\end{theorem} 
 
This Theorem shows the stability of our operating strategy. The conditionally independence between the two market demands actually covers a very wide range of real-life scenarios, and it is just a mild assumption. Even in the very common situation of demand correlation, the basic form of the optimal operating strategy remains unchanged. This means that our analysis results have laid a theoretical foundation for building a modernization chain, and can also be widely used to assist sophisticated managerial decisions.

\section{Conclusion}\label{sec6}
As enterprises carry out their production and selling activities on a global scale, they face the uncertainty and complexity of the global environment. Exploiting the difference timing of the selling season at different markets is an unique opportunity to increase the profitability of an enterprise. For enterprises with multiple production centers, the timing difference in the selling season allows secondary market production center to choose to purchase the remaining inventory in the primary market instead of producing.
 
Global markets provide enterprises with more production and selling opportunities, but at the cost of introducing more new uncertainties. Demand uncertainty is in many cases related to various demand characteristics, such as the geographic location of the market and consumers' tastes and evaluations of product quality. The primary market manager faces a longer planning cycle and needs to consider the demands of the secondary market, because the production decisions of the two markets are interrelated and the demands of the two markets may influence each other. In addition, demand risk will affect the quantity allocation of production and selling, while exchange rate risk will affect input costs and selling prices (Dornier et al. 1998). Therefore, it is interesting and important to provide enterprises with a stable operation strategy. This paper studies the above-mentioned problems by establishing a global robust newsvendor model in two situations: demand-related and demand-unrelated.
      
The global robust newsvendor model not only explains how the effective matching between supply and demand affects the choice of the enterprise's optimal strategy, but also manifests itself in the relatively insensitive to potential demand distribution and demand correlation. If there is very little information about market demand, the application of our model is particularly attractive to enterprises that are more averse to changes in their operational strategies. Besides, demand correlation does not affect the formulation and implementation of the optimal operational strategy. The production and selling allocation of each market is only related to the mean and variance of the corresponding market demand. This will be more conducive to the integration of enterprises into the process of globalization, and also provides a theoretical basis for how to build the traditional industrial chain and supply chain into a modern chain.
    
The simplicity, intuitiveness and efficiency of the optimal operation strategy presented in this paper are very attractive to enterprises. On the basis of considering the exchange rate risk, only the first and second moments of demand are required, which provides a great flexibility for the enterprise to approach a wider range of empirical sales data. This can better reflect actual business goals and decision-making behaviors, and also build a bridge to supplement the literature that combines transnational supply chain, Inventory transshipment and robust newsvendor problem. A further natural research question is whether the global operation strategy of an enterprise facing multiple risks still maintains symmetry, independence and demand irrelevance if the enterprise manager is risk-averse, or whether the operation strategy is more aggressive or more reserved compared with the risk-neutral one. In addition, it will be fruitful to further study the global robust operation strategy from the perspective of supply chain or competition.

\bibliographystyle{nonumber}

\end{document}